%% file: PaperForReview.tex
\documentclass[10pt,twocolumn,letterpaper]{article}

\usepackage[pagenumbers]{wacv} 

\usepackage{graphicx}
\usepackage{amsmath}
\usepackage{amssymb}
\usepackage{booktabs}
\usepackage{multirow} 
\usepackage{makecell}
\usepackage{amsmath,amssymb}
\usepackage{pifont}
\usepackage{bm}
\usepackage{subcaption}
\usepackage{color}
\usepackage{mathtools}
\usepackage[x11names]{xcolor}

\definecolor{dgreen}{rgb}{0.04,0.7,0.13}
\definecolor{maroon}{rgb}{0.75,0.07,0.03}
\definecolor{dblue}{rgb}{0.1,0.07,0.75}
\newcommand{\cmark}{{\color{dgreen}\ding{51}}}%
\newcommand{\cmarkred}{{\color{maroon}\ding{51}}}%
\newcommand{\xmark}{{\color{maroon}\ding{55}}}%
\newcommand{\xmarkgreen}{{\color{dgreen}\ding{55}}}%
\usepackage[pagebackref,breaklinks,colorlinks]{hyperref}

\input{preamble}

\usepackage[capitalize]{cleveref}
\crefname{section}{Sec.}{Secs.}
\Crefname{section}{Section}{Sections}
\Crefname{table}{Table}{Tables}
\crefname{table}{Tab.}{Tabs.}
\usepackage{float}

\begin{document}

\title{\OURS: A Point Cloud Diffusion Model of Human Shape in Loose Clothing}

\author{Siddharth Seth$^1$\thanks{work done as part of an internship at MPII.}\qquad Rishabh Dabral$^2$ \qquad Diogo Luvizon$^2$ \qquad Marc Habermann$^2$\\
Ming-Hsuan Yang$^1$ \qquad Christian Theobalt$^2$ \qquad Adam Kortylewski$^{2,3}$\\
$^1$UC Merced  ~~~~~ $^2$Max Planck Institute for Informatics ~~~~~ $^3$University of Freiburg\\
}

\maketitle

\input{sec/0_abstract}

\input{fig/fig_teaser_wacv}
\input{sec/1_intro}
\input{sec/2_related_works}
\input{sec/3_method}

\input{sec/3b_dataset}
\input{sec/4_experiments}
\input{sec/5_conclusion}

\clearpage
{\small
\bibliographystyle{ieee_fullname}
\bibliography{egbib}
}
\clearpage
\input{sec/suppl}
\end{document}

%% file: preamble.tex
\usepackage{placeins}
\usepackage{color}
\definecolor{darkorange}{rgb}{1.0, 0.55, 0.0}
\definecolor{darkgreen}{rgb}{0.0, 0.40, 0.0}
\definecolor{magenta}{rgb}{1.0, 0.0, 1.0}
\newcommand{\MH}[1]{}
\newcommand{\DL}[1]{}
\newcommand{\SID}[1]{}
\definecolor{RDcolor}{rgb}{0.6, 0.4, 0.8}

\newcommand{\EX}[1]{\mathbf{X}^{(#1)}}
\newcommand{\OURS}{PocoLoco\xspace} 

\usepackage{soul}



%% file: sec/0_abstract.tex
\begin{abstract}
Modeling a human avatar that can plausibly deform to articulations is an active area of research. We present \textit{\OURS} -- the first template-free, point-based, pose-conditioned generative model for 3D humans in loose clothing. We motivate our work by noting that most methods require a parametric model of the human body to ground pose-dependent deformations. Consequently, they are restricted to modeling clothing that is topologically similar to the naked body and do not extend well to loose clothing. 
The few methods that attempt to model loose clothing typically require either canonicalization or a UV-parameterization and need to address the challenging problem of explicitly estimating correspondences for the deforming clothes.
In this work, we formulate avatar clothing deformation as a conditional point-cloud generation task within the denoising diffusion framework.
Crucially, our framework operates directly on unordered point clouds, eliminating the need for a parametric model or a clothing template.
This also enables a variety of practical applications, such as point-cloud completion and pose-based editing -- important features for virtual human animation. 
As current datasets for human avatars in loose clothing are far too small for training diffusion models, we release a dataset of two subjects performing various poses in loose clothing with a total of 75K point clouds.
By contributing towards tackling the challenging task of effectively modeling loose clothing and expanding the available data for training these models, we aim to set the stage for further innovation in digital humans.
The source code is available at \url{https://github.com/sidsunny/pocoloco}. 
\end{abstract}

%% file: fig/fig_teaser_wacv.tex
\begin{figure}[tp]
        \begin{center}
	\includegraphics[width=\linewidth]{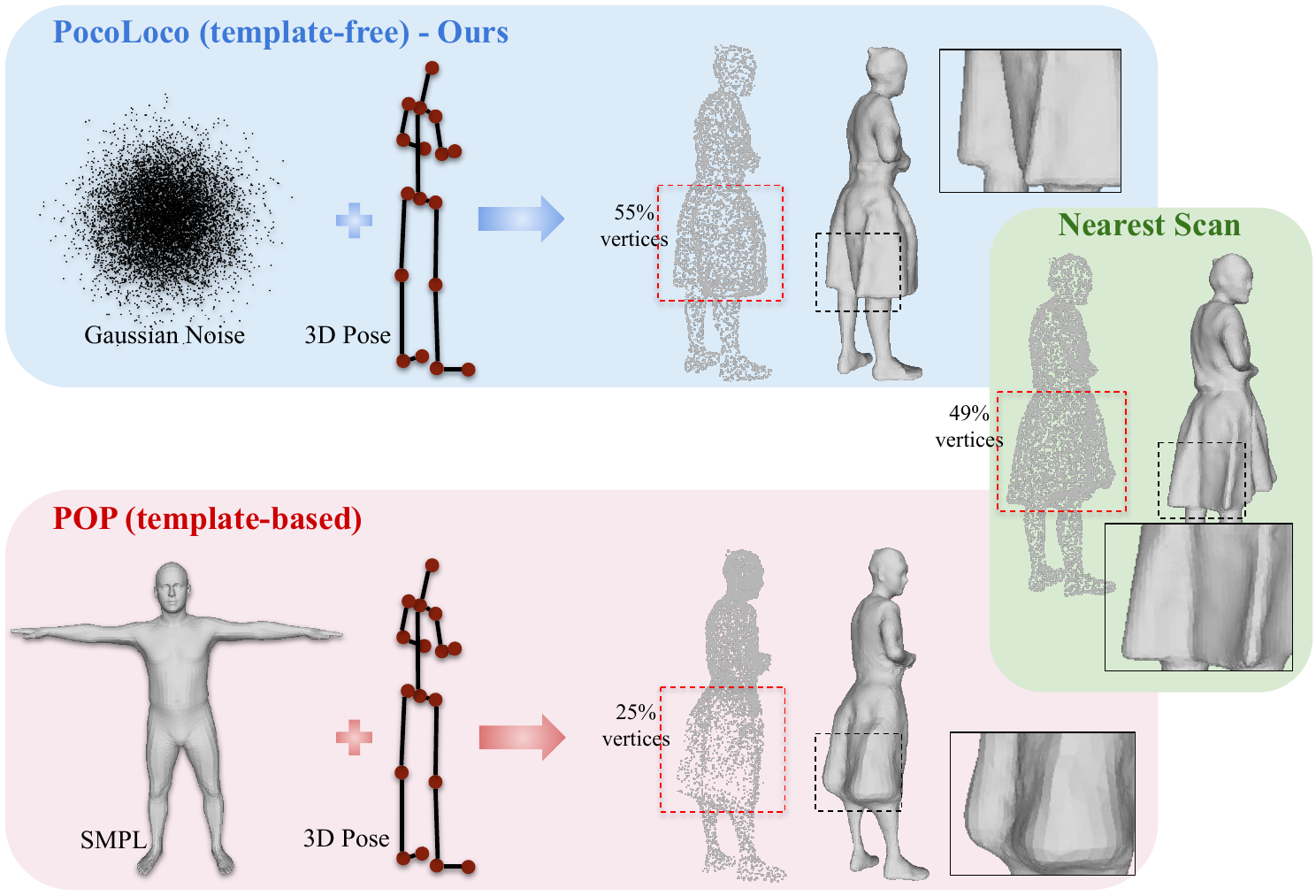}
	\caption
	{
	     Learning avatars in loose clothing with \OURS (Ours). Existing approaches, such as POP, rely on a parametric model to simulate clothing deformations. While these methods demonstrate promising results with body-fitted tight clothing, they often exhibit artifacts when modeling loose clothing, which differs topologically from the body shape. This issue is evident in the sparse points generated within the loose clothing regions (55\% vs 25\% vertices). Additionally, these approaches necessitate fitting a template to the input scan, which can be cumbersome and, from an artist's perspective, undesirable at times. In contrast, our learning-based approach, \OURS can model pose-dependent loose clothing deformations without requiring an underlying parametric body model, clothed templates, or complex linear-blend skinning. 
	}
	\label{fig:teaser}
        \end{center}
\end{figure}

%% file: sec/1_intro.tex
%
%
\section{Introduction} \label{sec:intro}
The tremendous recent advances in generative AI models are also driving a recent interest in creating articulated 3D avatars of clothed humans.
The ultimate goal is to create a personalized 3D model of a clothed body that can be controlled by the skeletal body pose, with the clothing deforming naturally as the body moves.
Such digital avatars should ideally be learned from just a given set of 3D meshes of a person performing various movements.
This will greatly simplify the creation of avatars for immersive experiences in applications like telepresence, virtual reality, content creation, and character animation, to name a few.

Creating human avatars is an artistic process and often requires several hours of work to produce plausible character animation.
In recent years, there have been tremendous improvements in automatic human avatar creation methods that generate avatars from inputs ranging from multi-view images~\cite{liu2021neural, zhao2022humannerf, peng2021neural} to highly-detailed 3D scans~\cite{POP:ICCV:2021,skirt2022}.
The majority of works base the animatable avatar on the parametric body model SMPL \cite{SMPL:2015} that provides coarse pose-dependent deformation and then either displace the vertices of the body template mesh \cite{burov2021dynamic,ma2020learning,POP:ICCV:2021,skirt2022} or use it to learn a neural implicit field that captures the clothing shape conditioned on the body parameters \cite{Saito:CVPR:2021,liu2021neural}.
However, approaches based on using SMPL as templates are typically limited to rather simple tight clothing topology as they need to learn an accurate surface correspondence from the template mesh to the 3D training scans \cite{Saito:CVPR:2021}.
This proves to be prohibitively simple and ineffective for loose clothing.
Consequently, even those methods that explicitly address modeling humans in loose clothing, but use SMPL as an underlying template, suffer from artifacts like tearing of the surface of loose clothing~\cite{POP:ICCV:2021}, or crease-formation between legs due to the UV parameterization~\cite{Saito:CVPR:2021}.

In this work, we introduce \OURS, the first template-free method based on a conditional diffusion model for learning articulated 3D avatars in loose clothing.
Our model operates directly on unordered point clouds without assuming point correspondences, thus avoiding the requirement of using parametric models or a clothing template (Fig.~\ref{fig:teaser}).
We use diffusion to learn the uncertainty associated with clothing deformation due to external forces. This is especially important when articulating the human body given a single pose which provides no information on the current action sequence and consequently the resulting deformations.
Concretely, given an input pose of the character, our goal is to produce a corresponding point cloud of the character that agrees with the pose with the clothing deforming naturally.

We formulate this conditional generative process within the generative framework of denoising diffusion models~\cite{dickstein,ddim,ddpm}.
However, extending diffusion models to 3D human point clouds is not a straightforward extension of 2D diffusion.
In Section~\ref{sec:meth_diff}, we discuss our various design choices behind the proposed transformer-based architecture for pose-conditioned human shape synthesis.
We highlight the importance of an appropriate scheduling policy for our setting, observing that in contrast to prior work \cite{zhou20213d} a quartic beta-schedule offers the best trade-off between utilization of the network capacity and high-quality shape synthesis with 10K points.
Notably, our model learns a person-specific distribution of cloth deformations without any explicit cloth deformation model from a dataset of point clouds that capture a few minutes of a person performing various movements.
Once learned, the avatar can be driven by arbitrary skeletal poses, akin to Linear Blend Skinning (LBS) but without having to learn it. 
Moreover, our generative approach allows us to model clothing deformations in a \textit{stochastic} manner, thus permitting us to generate multiple diverse hypotheses for the same input pose.
\par
As existing datasets for human avatars in loose clothing are too small
for training diffusion models, we release a dataset of two subjects in loose clothing with a total of  75K frames performing a variety of articulated motions.
We record this data in a multi-view camera setting and obtain high-quality 3D shape reconstructions following neural surface reconstruction proposed in \cite{wang2022neus2}.
Using our recorded data, we show that compared to popular methods for human modeling like NeuralActor \cite{liu2021neural}  we achieve a significant improvement in the representation of loose clothing. Even when compared to methods that focus on posing a pre-scanned high-quality template of the human subject, like DDC \cite{habermann2021}, our method performs competitively.
The main contributions of this work are three-fold:
\begin{itemize}
    \item We present \OURS, which to the best of our knowledge, is the first 1) template-free and 2) conditional generative model for articulated 3D avatars in loose clothing. Our model operates directly on unordered point clouds without assuming point correspondences.
    \item Going beyond 3D shape generation, our diffusion-based formulation also enables capturing the inherent stochasticity in the deformations of loose clothes, enabling several practical applications.
    \item To evaluate and benchmark our method, we capture a dataset of high-quality 3D reconstructions of $2$ subjects with 75K frames performing a variety of actions. 
    By expanding the available data for training large-scale generative models of humans in loose clothing, we are setting the stage for even more innovative breakthroughs in the future.
\end{itemize}

%% file: sec/2_related_works.tex
\input{tab/tab_related}

\vspace{-4mm}
\section{Related Work} \label{sec:related}
Learning deformable human models has been extensively studied in the 
literature, specially when considering template-based tracking~\cite{habermann2021, habermann2020deepcap, xu2018monoperfcap} and parametric human models~\cite{SMPL:2015, SMPL-X:2019, ghum}.
Fewer works have proposed completely template-free modeling, as discussed next. In what follows, we also review the most relevant works on diffusion models for 3D generation and existing approaches to modeling clothing deformation. We compare the characteristics of the most relevant existing works with our method in \cref{tab:related}.

\vspace{1mm}
\noindent \textbf{Template-free Human Models in Loose Clothes.}
To break free from the constraints of following a fixed topology per garment \cite{guan2012drape,habermann2020deepcap,pons2017clothcap,patel20tailornet}, so-called ``template-free" methods rely on an underlying SMPL \cite{SMPL:2015} body shape and deform that into a variety of clothing styles \cite{chen2021snarf,palafox2021npms,Saito:CVPR:2021,wang2021metaavatar,hu2024gaussianavatar}.
In particular, SCANimate~\cite{Saito:CVPR:2021} fits an SMPL model to 3D scans to obtain body joints and skinning weights. They propose to learn mappings to canonicalize the input scan and then repose it to get the input pose while following a cycle consistency.
Although such representations adapt to varied topologies, these methods do not perform well when modeling loose clothing. 
The problem originates from the need to canonicalize the training data, as analyzed in detail in \cite{skirt2022}. 
In contrast, point clouds provide an alternative representation. 
\cite{POP:ICCV:2021} predict a displacement field on the unclothed body based on the local pose and geometric features. 
Although it can animate a single scan with unseen poses and clothing, and handle pose-dependent deformations, it still suffers from seaming artifacts in skirts and loose clothing due to its reliance on the unclothed body model for local coordinates.
In a follow-up approach~\cite{skirt2022}, point clouds are obtained from scans and model clothing deformation in a local coordinate system without the need for explicit canonicalization.
However, these methods are still dependent on learning LBS weights which work well for tight clothing as they adhere to the body articulation constraints, unlike free-flowing loose clothing.
This is evident from the sparse set of points produced in the skirt and dress outfits~\cite{skirt2022}. ~\cite{prokudin2023dynamic} sidestep LBS and aim to model dynamics by learning deformation fields over a canonical frame. However, they still need point correspondences to model clothing deformation.

Besides point-based approaches,~\cite{Ianina2022BodyMapLF, 10208428} use dense correspondences between images of clothed humans and the surface of a 3D template to represent human avatars under varying clothing styles. Furthermore, ~\cite{Chen2023GMNeRFLG, 10.1007/978-3-031-19784-0_11, arnabdey, xiu2022icon} either use implicit-based approaches or combine with SMPL to exploit human body priors for learning 3D clothed avatars.

\OURS is the first completely template-free approach that neither depends on an underlying SMPL body mesh nor requires any LBS weights, hence making it easier to model non-rigid deformations in loose clothing.

\vspace{1mm}
\noindent \textbf{Diffusion Models for 3D Generation.}
Denoising diffusion probabilistic models~\cite{dickstein,ddim,ddpm} have recently become the go-to frameworks for generative modeling.
While there has been a deep interest in building generative models for a plethora of research problems pertaining to 2D images, the 3D domain is fundamentally more challenging and has only started receiving attention recently.
\cite{zhou20213d} use a point voxel representation based on PointNet++~\cite{qi2017pointnet++} to model 3D shape generation.
by adding noise to the input 3D point cloud and learning a distribution of added noise at each time step.
LION~\cite{zeng2022lion} on the other hand is designed as a hierarchical point cloud VAE~\cite{kingma2013auto}, denoising over the shape latent and latent point distributions.
While these methods model rigid objects, our work focuses on modeling the complex human pose articulation and non-linear clothing deformation.
%

%% file: tab/tab_related.tex
\begin{table}[tp]
\begin{center}
\scriptsize
\begin{tabular}{ccccc}
    \toprule

    \multirow{2}{*}{Methods} & \makecell{Scanned} & \makecell{Registration/} &  \makecell{Unseen} & \makecell{Generative}\\

    & \makecell{Template} & \makecell{SMPL} &  \makecell{Poses} & \makecell{Ability} \\

    \midrule

    DDC~\cite{habermann2021}	&              \cmarkred          &  \xmarkgreen    &      \cmark      &  \xmark  \\

    NeuralActor~\cite{liu2021neural}  &	      \xmarkgreen          &  \cmarkred    &     \cmark      &  \xmark     \\ 

    SCANimate~\cite{Saito:CVPR:2021}	 &    \xmarkgreen     & \cmarkred    &    \xmark      &   \xmark    \\

    POP~\cite{POP:ICCV:2021}	&     \xmarkgreen    &   
    \cmarkred     &     \xmark    &   \xmark  \\

    SkiRT~\cite{skirt2022}	  &                  \xmarkgreen    &  \cmarkred                &     \cmark   &  \xmark  \\ 

    \textbf{\OURS}	     &   \xmarkgreen   &   \xmarkgreen                &   \cmark           & \cmark \\

    %
    \bottomrule
    \end{tabular}
    \end{center}
    \caption
    {
    Characteristics of SOTAs on human modeling in loose clothing. \OURS does not require an underlying template, SMPL body shape, or LBS. Furthermore, it is the first generative method to model the ambiguity associated with loose clothing deformation. Green denotes a positive (and red a negative) trait. 
    }
    \label{tab:related}	
 \vspace{-4mm}
\end{table}

%% file: sec/3_method.tex
%
%
\section{Method} \label{sec:method}
Given the 3D keypoints, $K = \{k_1, k_2, \cdots, k_J\}$ of a clothed human, our goal is to construct a 3D point-cloud, $X = \{x_1, x_2, \cdots, x_N\}$ of the person justifying the given pose and the corresponding pose-dependent clothing deformations.
We pose this as a conditional generation problem, wherein the task is to learn the conditional distribution of point clouds, $p(X|K)$, given the keypoints. 
To this end, we leverage the ability of Denoising Diffusion Probabilistic Models (DDPM)~\cite{ddpm} to model this conditional distribution.
While existing methods~\cite{zeng2022lion, liu2019point} have previously demonstrated the ability of diffusion models in generating point-clouds of rigid objects in the ShapeNet dataset~\cite{shapenet2015}, it is challenging to extend them to human point-clouds due to two reasons.
First, humans exhibit high degrees of articulation, which induces pose-dependent deformation of the shape and is thus challenging to model.
Second, modeling the surface deformations becomes significantly more challenging if the subjects wear loose clothing.
These reasons make a simple extension of the aforementioned methods into a human-generation setting infeasible.%
In the following, we introduce DDPMs for human generation (\cref{sec:meth_diff}), network design (\cref{sec:network}), and the training strategy (\cref{sec:training}).

\subsection{Diffusion for Human Generation}\label{sec:meth_diff}
As mentioned above, we formulate human point-cloud generation as a conditional generation task.
Using a denoising-diffusion framework, we generate the point cloud $\mathbf{X}$ by progressively denoising a random noise vector, $\mathbf{z} \in \mathbb{R}^{3N}$.
In the forward diffusion process, the ground-truth point-cloud, $\EX{0}$, is repeatedly corrupted by adding random Gaussian noise, $\mathbf{\epsilon}$, for $T$ timesteps:
\begin{equation}
    q \big( \EX{1:T} | \EX{0} \big) = \prod_{t=1}^{T} q\big(\EX{t} | \EX{t-1}\big)
\end{equation}
The diffusion kernel $q(\EX{t} | \EX{t-1}) = \mathcal{N}(\EX{t}| (1-\beta_t)\EX{t-1}, \beta_t\mathbf{I})$ adds the Gaussian noise at every step in a Markovian fashion and $\beta_t$ is a hyperparameter controlling the rate of diffusion.
With sufficiently large $T$, $X^{(T)}$ is expected to follow the normal distribution $\mathcal{N}(0, \mathcal{I})$.
As~\cite{ddpm} show, one can simply compute $X^{(t)}$ at timestep $t$ using:
\begin{equation}\label{eq:ddpm_equation}
    \EX{t} = \sqrt{\bar{\alpha}} \EX{0} + \sqrt{1 - \bar{\alpha}}\mathbf{\epsilon}
\end{equation}
%

\input{fig/fig_overview}
\par
To sample a point cloud in the distribution of 3D point clouds, we need the following denoising function~\cite{dickstein}:
\begin{equation}
    p(\EX{0:T}) = p(\EX{T}) \prod_{t=1}^T p(\EX{t-1} | \EX{t})
\end{equation}
where $\EX{T} = \mathbf{z}$ is the initial random noise vector to sample from.
However, computing $p(\EX{t-1} | \EX{t})$ is intractable and is, therefore, approximated using a neural network,  $f_{\theta}(\EX{t-1} | \EX{t}, t)$, that can be trained to learn this function.
Instead of directly learning to predict $\EX{t-1}$ from $\EX{t}$,~\cite{ddpm} show that one can also make the network predict the original noise $\mathbf{\epsilon}$ used in~\cref{eq:ddpm_equation}. Specifically, we parametrize the function $f_{\theta}(\EX{t-1} | \EX{t}, t)$
using a point-transformer \cite{nichol2022pointe} that takes as input the noisy point-cloud at $\EX{t}$ at diffusion step $t$ and estimates the noise $\mathbf{\epsilon}$ that was used to corrupt the original point-cloud $\EX{0}$.
The point-transformer consists of several layers of multi-head self-attention blocks \cite{vaswani2017attention}.

\begin{figure}[tp]
    \centering
    \includegraphics[height=4cm,width=.99\linewidth]{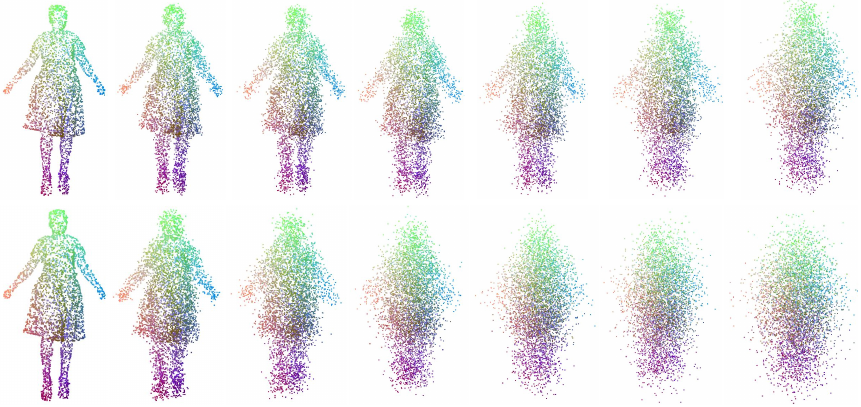}
    \caption{Comparison of our $\beta$ varying schedule (top) \textit{vs} the standard linear schedule (bottom) for the first 300 (of 1000) steps in the diffusion process. The human shape is preserved longer throughout the diffusion process, which facilitates the learning process and significantly improves the performance  (\cref{sec:ablation}). }
    \label{fig:forward_diffusion}
\end{figure}

\subsection{Pose Conditioning Mechanism} \label{sec:network}
Similar to other types of generative
models~\cite{mirza2014conditional}, diffusion models
are in principle capable of modeling conditional distributions of the form $p(z \vert c)$.
In the context of point cloud synthesis, combining the generative power of diffusion models with other types of conditionings beyond class-labels~\cite{zhou20213d} is so far an under-explored research area.

We turn point-based diffusion models into more flexible conditional generators by augmenting the underlying Transformer backbone with the cross-attention mechanism~\cite{vaswani2017attention}, which has proven effective as a conditioning mechanism in image diffusion models~\cite{rombach2022high}. 
Fig.~\ref{fig:overview} presents the overall schema of the network design. 

To that end, we first positionally-encode \cite{mildenhall2021nerf} each 3D keypoint $k \in K$ into a $2L$-dimensional space using: 
\[
    c = \oplus_{k\in K}(\sin(2^0\pi k), \cdots, \sin(2^{L-1}\pi k), \cos(2^{L-1}\pi k))
\]
This positionally encoded set of 3D keypoints representing the joint positions of the character can be introduced to the denoising function $f_{\theta}(\EX{t}, t, c)$ as a conditioning signal.
Specifically, we formulate the cross-attention operation on the intermediate features $\mathrm{x}$ of $\EX{t}$ as:
\begin{align}
    Attn(\EX{t}, \mathbf{c}) = \operatorname{softmax}\big( \frac{{Q} \otimes {K}^{\top}}{\sqrt{d}} {V}\big), \\
    Q = W_q\mathrm{x}, \quad K =  W_k\mathbf{c}, \quad V = W_v\mathbf{c}
\end{align}
where $W_q$, $W_k$, and $W_v$ are linear matrices projecting the intermediate features as the query, key, and value features, respectively. 
In the case of self-attention, one can simply substitute $c$ with $\mathrm{x}$.

\subsection{Training} \label{sec:training}
In the following, we provide some of the key steps of the derivation of the loss function to train our diffusion model and refer to~\cite{ddpm} for more details about the full derivation.
We derive the training objective from a variational upper bound on the Negative Log-Likelihood (NLL). 
Obtaining the upper bound requires we specify a surrogate distribution referred to as $q$ which governs the diffusion process. 
Following Jensen's inequality, the NLL of a data point $\EX{0}$ can be upper-bounded using $q$:
\begin{equation}
    -\log p_\theta(\EX{0})\leq\mathbb 
    E_q\left[-\log\frac{p_\theta(\EX{0:T})}
    {q(\EX{1:T}|\EX{0})}\right]
    \coloneqq
    \mathcal{L}(\EX{0}|\theta)\,,
\end{equation}
where $\EX{t_1:t_2}$ is the set $(\EX{t_1},\ldots,\EX{t_2})$. 
The loss $\mathcal{L}(\EX{0}|\theta)$ which upper bounds the NLL takes a simple and intuitive form:
\begin{align}
    \mathcal{L} &= \mathbb{E}_{t\sim[1,T]} \big[ \vert\vert\mathbf{\epsilon} - f_{\theta}(\EX{t}, t, c)\vert \vert^2 \big]\\
                &= \mathbb{E}_{t\sim[1,T]} \big[ \vert\vert\mathbf{\epsilon} - f_{\theta}(\sqrt{\bar{\alpha}}\EX{0} + \sqrt{1 - \bar{\alpha}\epsilon}, t, c)\vert \vert^2 \big]
\end{align}
Hence the denoising network $f_{\theta}(\EX{t}, t, c)$ can be trained to estimate the input noise $\epsilon$, at each time step $t$. \\

\input{fig/fig_dataset}
\noindent \textbf{Scheduling Policy}: In our experiments, we noted that linear scheduling can be wasteful, as it quickly diffuses into a random point cloud, thereby, \textit{wasting} significantly many diffusion steps.
Ideally, one would want the diffusion process to be spread out across several time steps.
We found that a quartic $\beta$ schedule works best for our setting $\beta(t)=\left(\frac{t}{10000}\right)^4$, where $t=[1,\dots,1000]$ is the number of steps in the diffusion process.
\cref{fig:forward_diffusion} demonstrates the effect of our $\beta$-varying schedule over the popular linear schedule
by visualizing every 100th step in the diffusion process.
Note the significantly reduced number of time steps where the point cloud is completely random.

%% file: fig/fig_overview.tex
%
%
\begin{figure}[tp]
	\centering
	\includegraphics[width=.99\linewidth]{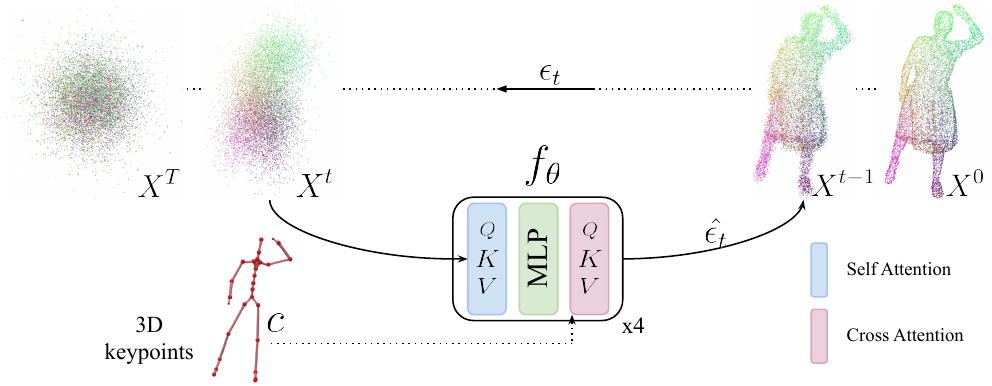}
	\caption
	{
	    Method overview. We visualize the diffusion process (from right to left) incrementally adding noise $\epsilon_t$ at every diffusion step $t$ to the point cloud of a human in loose clothing. During the reverse (generative) process, Gaussian noise is sampled $X^T$ and noise is progressively removed by predicting a residual noise $\hat{\epsilon}_t=f_\theta(X_t)$. The diffusion model $f_\theta$ is composed of cross- and self-attention layers with query, key, and value tokens and a multi-layer perceptron (MLP). The skeletal pose-conditioning is applied in the cross-attention layer.
	}
	\label{fig:overview}
\end{figure}
%
%

%% file: fig/fig_dataset.tex
%
%
\begin{figure}[tp]
	\centering
	\includegraphics[width=.99\linewidth]{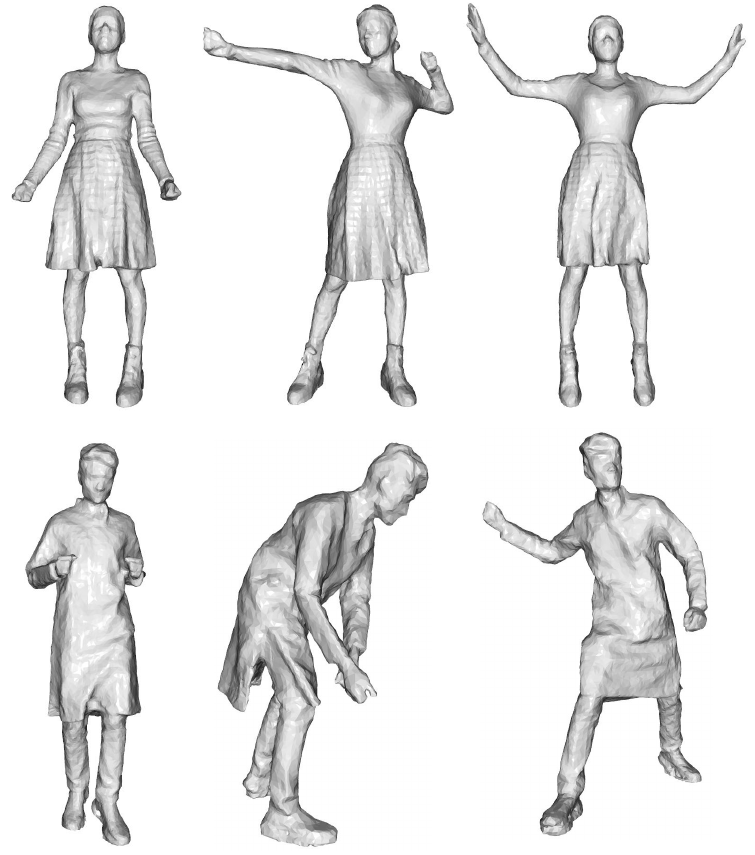}

	\caption
	{
	    Visualization of samples from our LOOSE dataset with two subjects performing
various poses in loose clothing with a total of 75K point clouds. The data was recorded in a multi-view camera setting and the 3D reconstructions were obtained following the approach as proposed in Neus2 \cite{wang2022neus2}. 
     We note that our recorded data has a resolution of approximately 40K points, but here we visualize it with 10K points, as this corresponds to the data as used for training our model. 
	}
	\label{fig:dataset}
     \vspace{-5mm}
\end{figure}
%
%

%% file: sec/3b_dataset.tex
\section{Dataset for Humans in Loose Clothing} 
\label{sec:datasets}
Current datasets of humans in loose clothing are vastly limited in terms of size. 
For instance the popular ReSynth \cite{POP:ICCV:2021} data offers three subjects wearing distinguished loose outfits, but the standard train and test split consist only of $984$ and $347$ frames respectively per outfit.
We found that this small amount of data is not sufficient for large-scale machine learning approaches such as our model.

One of the contributions of this work is that we expand the available data for
training large-scale machine learning approaches for generative human modeling in loose clothing, setting the stage for even more innovative
breakthroughs in the future.
In particular, we created a new dataset, called \textit{LOOSE}, which consists of 
    two subjects each both with loose clothing, one wearing a long loose shirt and the other a skirt (see \cref{fig:dataset}). 
Each sequence is recorded at 25fps and is split into a training and testing recording which contain around 30,000 and 10,000 frames respectively. 
 The training and test motions are significantly different from each other. Following common practice, we acquired separate recordings for training and testing (instead of randomly sampling from a single sequence). For each sequence, we asked the subject to perform a wide range of motions like “dancing” which was freely interpreted by the subject. We recorded with 50 to 101 synchronized and calibrated cameras at a resolution of 1285 × 940. 
 Subsequently, we obtain high-quality 3D shape reconstructions following the neural surface reconstruction approach proposed in \cite{wang2022neus2}, giving very high-quality reconstruction of loose clothing.
 We will release the new dataset, as no other image sets at that scale with human subjects in loose clothing exist for evaluation. 

%% file: sec/4_experiments.tex
%

\input{fig/fig_smpl_pop_pocoloco}

\vspace{-3mm}
\section{Experimental Results} 

\label{sec:results}
We demonstrate the merits of the proposed method against state-of-the-art approaches with ablation studies.  

\vspace{-1mm}
\subsection{Evaluation Details}
\vspace{-1mm}

\textbf{Datasets.} In addition to our proposed large-scale LOOSE dataset, we use the publicly available DynaCap dataset \cite{habermann2021real} and evaluate two subjects. One is wearing a tight shirt and shorts, and the other is wearing loose clothing - a top and a skirt. We use the official training and testing split which contain around 20,000 and 7000 frames respectively. 

\vspace{1mm}
\noindent \textbf{Baselines.} We compare \OURS with a number of state-of-the-art human modeling methods, such as the template-free model NeuralActor \cite{liu2021neural} and the template-based model DDC \cite{habermann2021}, that reported results on the DynaCAP dataset. We also compare with POP~\cite{POP:ICCV:2021}, a point-based method that requires a SMPL template for predicting delta clothing deformation.
Finally, we perform extensive ablation studies of our
method in \cref{sec:ablation}.

\noindent \textbf{Network Architecture.} We use positional encoding for input and pose conditioning with $L=7$. To condition on the pose, we first feed it to a $512$ dimensional projection layer. The input is processed similarly to obtain per-point $512$ dimensional features. We use a Transformer \cite{vaswani2017attention} as the backbone architecture with $8$ layers and $4$ attention heads. Each head has a depth/attention channels of $128$. Each layer consists of self-attention on the input, followed by a cross-attention on the pose-conditioning. Outputs from each of the self and cross-attention layers are separately passed through MLPs with GELU \cite{hendrycks2016gaussian} activation.

\subsection{Qualitative Results} 
\vspace{-1mm}
\label{sec:qualitative}
\cref{fig:qualitative} shows qualitative results of our method on two subjects from the DynaCap dataset and the two subjects included in our LOOSE dataset.
We note that our method produces point clouds, and for visual convenience, we render the point clouds as meshes using the standard Poisson surface reconstruction \cite{kazhdan2006poisson} implemented in MeshLab (\url{https://www.meshlab.net}).
We can observe that our \OURS method generates faithful high-quality human avatars in loose clothing even in varying poses. 
Importantly, our model attends to diverse possibilities in clothing deformation in various poses with the deformations being realistic and naturally matching the overall body pose.
%
%

\input{fig/fig_qualitative_results}
\input{fig/fig_comparison}
\input{tab/tab_quantitative_image}

\subsection{Comparison to Prior Work} 
\vspace{-1mm}\label{sec:evaluation}
\vspace{-.1cm}
Similar to SoTA methods, we report results using Chamfer Distance metric (cm) as well as Scan-to-Model (cm) and Model-to-Scan distance (cm) averaged over all test samples for a subject. We summarize the quantitative results in \cref{tab:quantitativeimage}. Compared to NeuralActor, our \OURS achieves a compelling improvement, particularly for the subject in loose clothing. Moreover, even when compared to the template-based method DDC, which can be considered an upper bound in terms of performance as it makes much stronger assumptions about the training supervision, our \OURS performs competitively, even outperforming in the case of tight clothing while performing slightly worse on loose clothing. We also note that the reported numbers for our generative method depend on the randomly sampled input latent code. As different latent codes induce different folding patterns (\cref{sec:diversity}) our method has a slight disadvantage over the deterministic methods when evaluating quantitatively. Although we report our quantitative results as an average of ten runs, we hypothesize that using predictions closest to the ground truth meshes as used by other generative methods should boost our performance even further. Lastly, we also compare our method with POP which also follows a point setting, albeit, still using a template mesh as an added advantage. Although we note that POP has a better quantitative performance than ours, we show through qualitative results that the chamfer distance (CD) metric does not capture the task of clothing deformation as intended. Here, we draw the attention of the reader towards directly using SMPL fittings for evaluation. As shown in \cref{tab:quantitativeimage}, this already achieves a low CD, further establishing that CD alone is insufficient for modeling deformations in loose clothing.

\noindent \textbf{User study.} We compare PocoLoco against reconstructions from POP by randomly picking 40 poses from the DynaCap dataset. We ask the users to choose between the two methods the one with better pose and clothing geometry reconstruction. We collect ratings from 16 users with a total of 640 ratings. PocoLoco is chosen over POP in 85.6\% of the comparisons. Note that 50\% is the upper bound of any method for the comparisons where the choice is random.

\noindent \textbf{Qualitative Comparison.} We show a visual comparison of the results in \cref{fig:comparison}. Notably, NeuralActor produces visible artifacts in the shape reconstruction, whereas \OURS generates faithful human shapes even in challenging poses. The DDC model can generate high-frequency details, as these are captured in the pre-scanned template of the human subject. However, it also produces visible high-frequency noise. In contrast, our model generates genuine and smooth shapes, though missing some of the high-frequency details that are visible in the DDC generation.

\noindent \textbf{Performance comparison with POP and SMPL.} \cref{fig:smpl} shows how a SMPL mesh fitted to the ground truth scan and reposed to the test sequence fares against POP and PocoLoco. The first row illustrates that while there is a negligible difference in CD between POP and PocoLoco, POP largely fails to capture the clothing deformations while PocoLoco can model these with the uniformly distributed points across the body and clothing. The posed SMPL mesh lacks clothing and thus has a higher CD. The second row shows a more relevant case of clothing deformation not captured by CD. The posed SMPL mesh cannot represent any clothing deformation but only differs by 1.2 cm in CD from PocoLoco. On the contrary, there is a similar 1.1 cm difference in CD between PocoLoco and POP, despite both being much closer in terms of body and clothing geometry. This shows that while using CD is a good metric for measuring the overall reconstruction, it does not consider the smoothness introduced in the reconstructions thereby ignoring the important pose-dependent clothing deformations.

\noindent \textbf{Point sparsity in loose clothing.} As POP generates a displacement vector corresponding to the unclothed body using SMPL, this results in points being located near the body surface. We illustrate in \cref{fig:smpl}, that,   this in turn gives sparse points in loose clothing regions such as skirts which is undesirable (leading to artifacts) and does not necessarily translate to errors captured by the quantitative measures. POP produces 33\% and 27\% of all the vertices at the loose clothing region compared to 49\% and 46\% vertices by PocoLoco for the two poses shown. While POP can reconstruct the overall pose-dependent clothing, it loses details on clothing deformation and produces smooth regions. Our diffusion-based model exhibits a strong inductive bias for the clothed body, generating points uniformly across the body and cloth regions, with noticeable learned clothing deformations.

%
%

\input{fig/fig_diversity}
\input{tab/module_ablation}
\vspace{-1mm}
\subsection{Clothing Deformation Uncertainty} 
\vspace{-1mm}
\label{sec:diversity}
Our generative model for modeling pose-dependent deformation in loose clothing explicitly accounts for the uncertainty in clothing deformation. This is extremely useful as representing the clothing deformation for a specific pose becomes particularly challenging when only a single pose is provided. This is because the sources of various external forces exerted on the person during motion are unknown for a single pose. In ideal scenarios, motion cues such as the direction of a person's movement would be useful. Nevertheless, we show in \cref{fig:diversity} that our model can account for the different clothing folds when presented with a single pose.
We visualize this by denoising different random input noises for the same pose conditioning
In comparison, prior arts always output a deterministic single clothing deformation when generating results for the same pose.
%


%
%
%
%
%
%
%
%
%
%
%
%
%
%
%
%
%
%

%
\subsection{Ablation} 
\vspace{-1mm}
\label{sec:ablation}
\vspace{-.1cm}
We perform an extensive ablation on our architecture type, depth, and beta scheduling (see \cref{fig:module_ablation}).
\noindent \textbf{Cross, Self, Cross+Self Attention.} We perform studies on the efficacy of using only self, only cross, and a combination of self and cross attention in the proposed Transformer architecture. Though self-attention works reasonably well, it takes a longer time to converge. Cross attention on the other hand does not converge after a long training time but can reasonably model the pose. A combination of self and cross-attention brings the best of both worlds by attending to the conditioned pose and faster convergence.

\noindent \textbf{Number of layers in Transformer.} We experiment with 2, 4, and 8 layers and see that the performance plateaus at 8 layers. The best result is obtained with 8 layers, but with a smaller improvement compared to the difference between 2 and 4 layers, indicating a saturation.

\input{fig/fig_applications}

\noindent \textbf{Linear vs Quartic beta scheduling.} Beta scheduling plays an important role in learning diffusion models. We observe that linear scheduling adds noise too quickly to the input samples, destroying the details early on. Instead, the quartic scheduling retains the details for more steps and smoothly converts the input samples to a random noise thereby giving the model more time to recover the details.

\subsection{Applications}
\vspace{-1mm}
\label{sec:apps}
Our simple yet effective clothed body representation in the form of point cloud, coupled with the power of diffusion models enables a variety of applications, such as point cloud completion and pose editing. We show these results in \cref{fig:applications}. Point cloud completion may help in filling holes obtained, for example, during the data acquisition process. On the other hand minor changes in pose can easily be attended to using a few steps of diffusion, resulting in smooth pose editing. We also show that our method can faithfully reconstruct pose-dependent clothing on OOD poses.


%% file: fig/fig_smpl_pop_pocoloco.tex
\begin{figure*}[t]
	\centering
	\includegraphics[width=.99
 \linewidth]{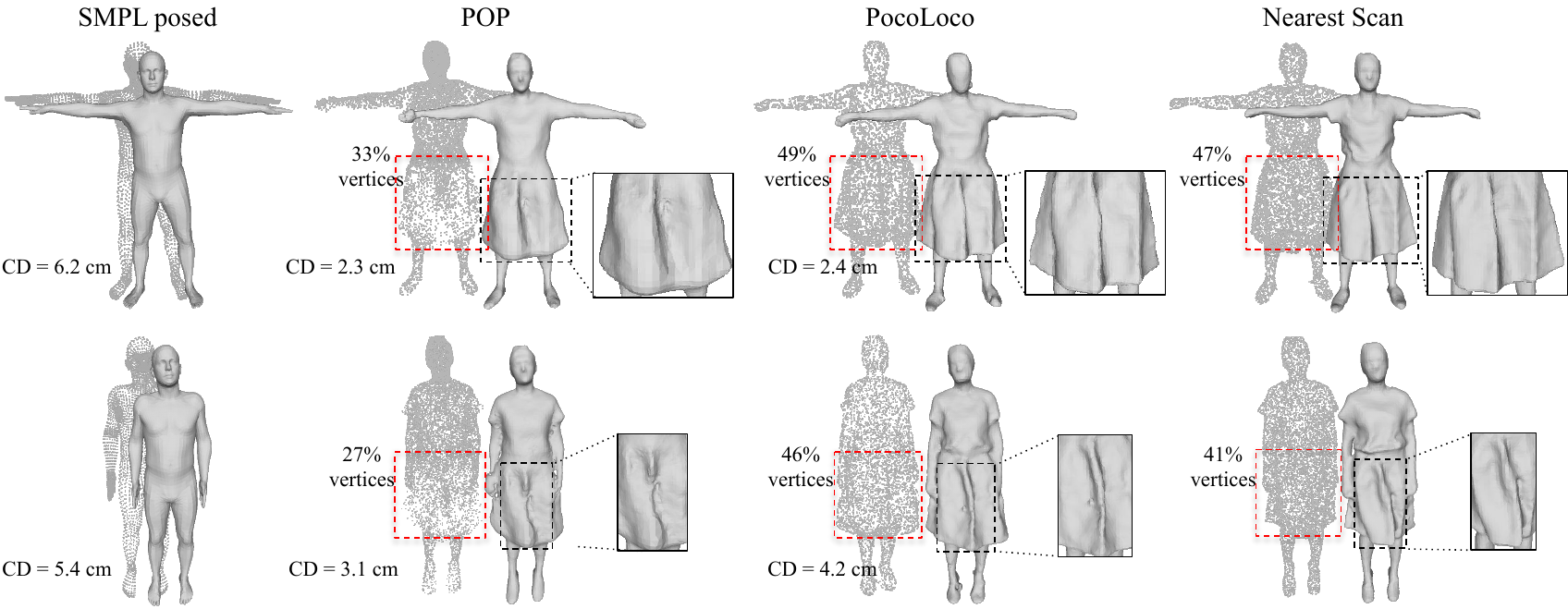}
	\caption
	{
	    Qualitative comparison of using a posed SMPL mesh as opposed to POP and PocoLoco. We show that the Chamfer Distance metric does not give a true representation of modeling clothing deformation.
	}
	\label{fig:smpl}
        \vspace{-4mm}
\end{figure*}

%% file: fig/fig_qualitative_results.tex
%
%
\begin{figure}[tp]
	\centering
	\includegraphics[width=.99\linewidth]{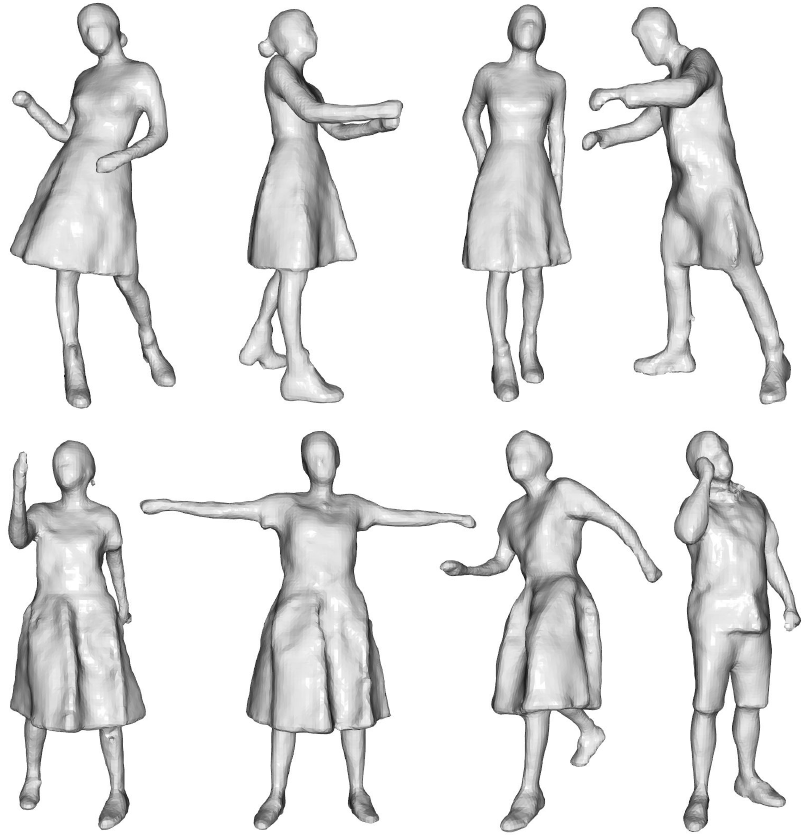}
	\caption
	{
	     Qualitative results of avatar generation with our \OURS model. We visualize models for four subjects in various challenging poses. The first shows two subjects in loose clothing, and the bottom row shows the first three samples of a subject in loose clothing and one subject in tight clothing (right).
	}
	\label{fig:qualitative}
        \vspace{-7mm}
\end{figure}
%
%

%% file: fig/fig_comparison.tex
%
%
\begin{figure}[tp]
	\centering
	\includegraphics[width=.95\linewidth]{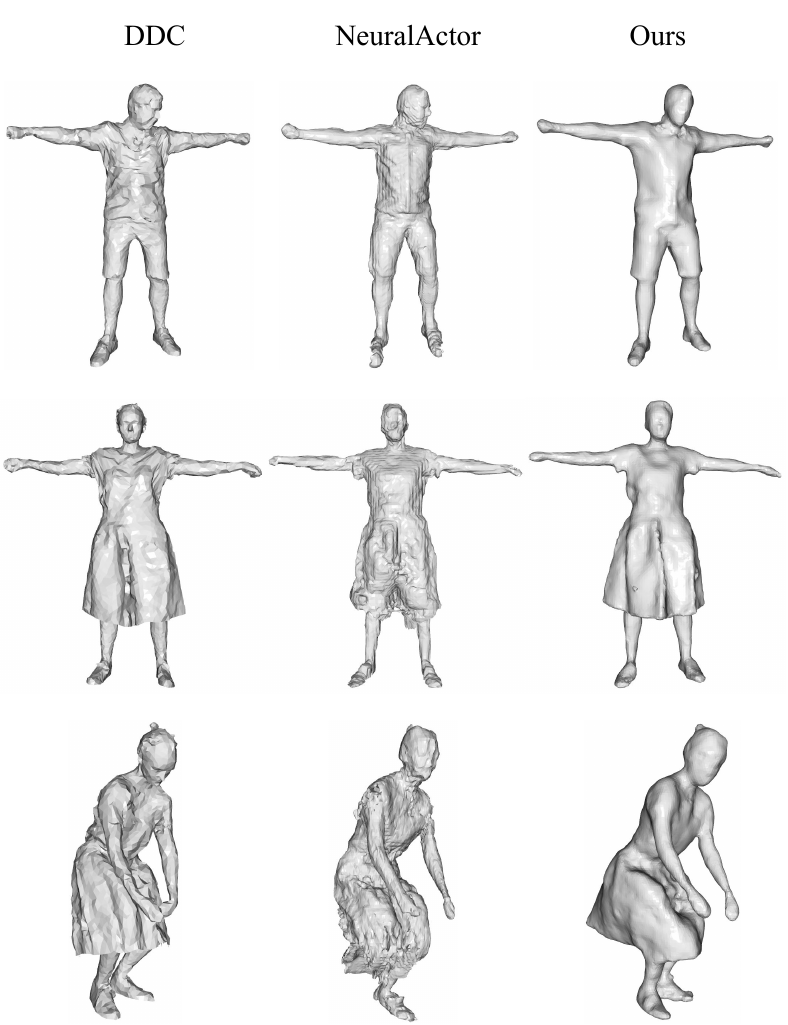}
	\caption
	{
	     Comparison to NeuralActor and DDC showing results for one subject in tight (top) and one in loose clothing (row 2, 3).
	}
	\label{fig:comparison}
\end{figure}
%
%

%% file: tab/tab_quantitative_image.tex
\begin{table}[tp]
	\footnotesize
	\centering
	 \setlength\tabcolsep{4.0pt}
  \scriptsize
    \begin{tabular}{lcccccc}
        \multicolumn{7}{c}{DynaCap}\\
    \toprule
 		  &  \multicolumn{3}{c}{Tight Clothing} & \multicolumn{3}{c}{Loose Clothing}\\

		\cmidrule{2-7}
        &Cham$\downarrow$&M2S$\downarrow$&S2M$\downarrow$&Cham$\downarrow$&M2S$\downarrow$&S2M$\downarrow$\\
		\midrule
          DDC 
		 & 2.93 & 1.39 & 1.54 &  4.06 & 1.89 & 2.17 \\
		 \midrule
          NeuralActor 
		 & 3.17 & 2.15 & 1.02 & 5.46 & 3.30 & 2.16 \\
		 
         SMPL fitting & - & - & - &  6.4 & 2.27 & 4.13  \\
         POP 
		 & - & - & - &  3.27 & 1.28 & 1.99  \\
   
        Ours & 2.48 & 1.22 & 1.26 & 4.22 & 2.12 & 2.1 \\
                   \bottomrule
    \end{tabular}
    \\
    
    \vspace{.25cm}
    \scriptsize
    \begin{tabular}{lcccccc}

        \multicolumn{7}{c}{LOOSE}\\
        \toprule
        &
        \multicolumn{3}{c}{Subject 1} & \multicolumn{3}{c}{Subject 2 (Skirt)}\\
        \cmidrule{2-7}
        &Cham$\downarrow$&M2S$\downarrow$&S2M$\downarrow$&Cham$\downarrow$&M2S$\downarrow$&S2M$\downarrow$\\
        \midrule
        
                  Ours & 2.87 & 1.39 & 1.48& 3.63 & 1.92 & 1.71  \\
                  \bottomrule
    \end{tabular}
    \\
 \vspace{.15cm}
        	\caption{ 
	Quantitative results reporting chamfer distance ($\downarrow$), scan-to-model and model-to-scan distance in cm. We first compare our method with template-based methods on the DynaCap dataset to  NeuralActor~\cite{liu2021neural} and DDC~\cite{habermann2021}. Next, we compare it with the SoTA point-based method POP~\cite{POP:ICCV:2021}. Finally, we report our results on the proposed LOOSE dataset. All results are for 10K points.
	}
	\label{tab:quantitativeimage}
    \vspace{-4mm}
\end{table}

%% file: fig/fig_diversity.tex
%
%
\begin{figure}[tp]
	\centering
	\includegraphics[width=.95
 \linewidth]{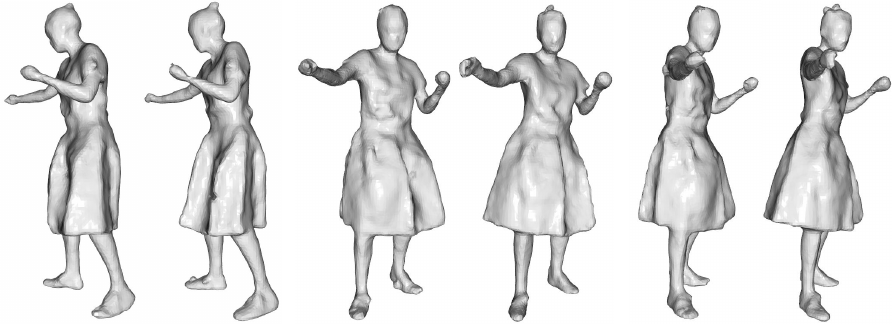}
	\caption
	{
	    Visualization of modeling uncertainty in clothing deformation when sampling an avatar with variable input noise for the same pose. Note the variation in skirt deformations.
	}
	\label{fig:diversity}

\end{figure}
%
%

%% file: tab/module_ablation.tex
\begin{figure}[tp]
    \begin{subfigure}{0.32\linewidth}
        \centering
        \scriptsize
        \begin{tabular}{cc}
            \toprule
            \textbf{Modules}  & \textbf{CD}$\downarrow$ \\
            \midrule
            Cross             & 5.04 \\
            Self              & 4.72 \\
            Self + Cross      & 4.22 \\
            \bottomrule
        \end{tabular}
        \caption{Modules}
    \end{subfigure}%
    \begin{subfigure}{0.37\linewidth}
        \centering
         \scriptsize
        \begin{tabular}{cc}
            \toprule
            \textbf{\# Layers} & \textbf{CD}$\downarrow$ \\
            \midrule
            2 x (Self + Cross) & 4.65 \\
            4 x (Self + Cross) & 4.54 \\
            8 x (Self + Cross) & 4.22 \\
            \bottomrule
        \end{tabular}
        \caption{\# Layers}
    \end{subfigure}%
    \begin{subfigure}{0.31\linewidth}
        \centering
         \scriptsize
        \begin{tabular}{cc}
            \toprule
            \multirow{2}{*}{\textbf{Scheduling}} & \multirow{2}{*}{\textbf{CD}$\downarrow$} \\
                   & \\
            \midrule
            Linear & 5.32 \\
            Quad   & 4.22 \\
            \bottomrule
        \end{tabular}
        \caption{Scheduling Policy}
    \end{subfigure}
    \caption{Ablation on the structure of the attention modules, the number of layers in the model, and the $\beta$ scheduling policy. We note the importance of all our design choices on the performance of our method, particularly the importance of the attention architecture and the scheduling policy.}
    \label{fig:module_ablation}
    \vspace{-4mm}
\end{figure}

%% file: fig/fig_applications.tex
\begin{figure}[tp]
	\centering
	\includegraphics[width=.99
 \linewidth]{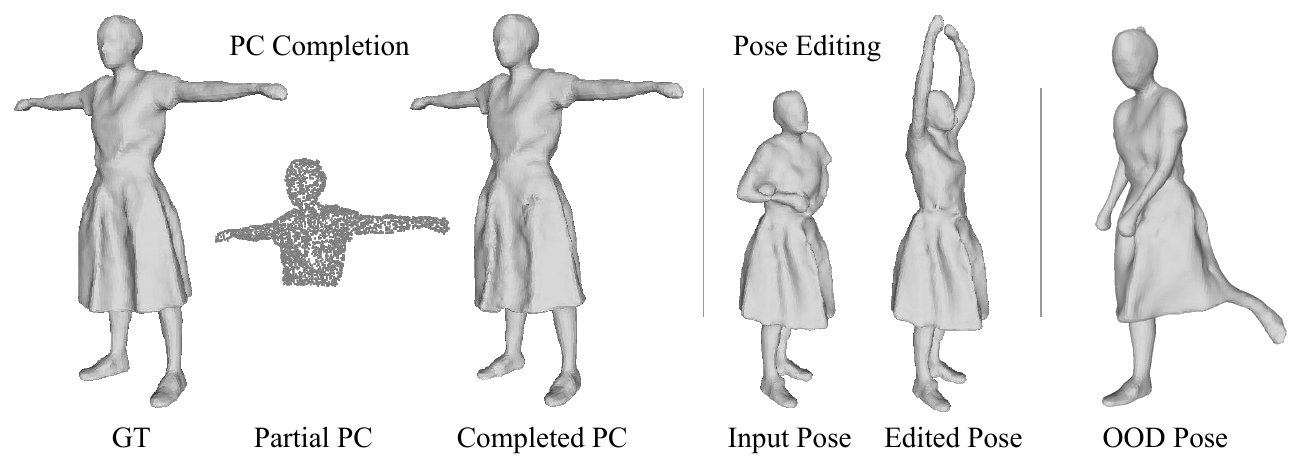}
	\caption
	{
	    PocoLoco can complete partial PCs, change the pose without affecting clothing deformations, and generalize to diverse poses, all while attending to challenging loose cloth deformations.
	}
	\label{fig:applications}
        \vspace{-5mm}
\end{figure}

%% file: sec/5_conclusion.tex
%
%

\section{Conclusion} \label{sec:conclusion}
\vspace{-2mm}
We present \textit{\OURS} -- a point-based, pose-conditioned generative model for 3D humans in loose clothing.
We achieve high-quality results, without assuming any registered clothing template, SMPL proxy shape, or linear-blend skinning, thereby following a completely template-free learning approach.
Our work makes an important step forward towards large-scale generative modeling of 3D human avatars, providing a significant improvement in the representation of loose clothing.
We also curate a dataset of high-quality point clouds of humans in loose clothing performing various poses, totaling 75K point clouds. By expanding the available data for training these models, we are setting the stage for further innovative breakthroughs in the space of digital humans.\\

\noindent \textbf{Acknowledgements.} AK acknowledges support via his Emmy Noether Research Group, funded by the German Research  Foundation (DFG) under Grant No. 468670075.

%% file: sec/suppl.tex
\appendix
\setcounter{page}{1}
\setcounter{section}{0}
\setcounter{table}{0}
\setcounter{figure}{0}

\onecolumn
\begin{center}
\LARGE\textbf{Appendix}
\end{center}

\begin{figure*}[!ht]
\centering
  \includegraphics[width=0.9\textwidth]{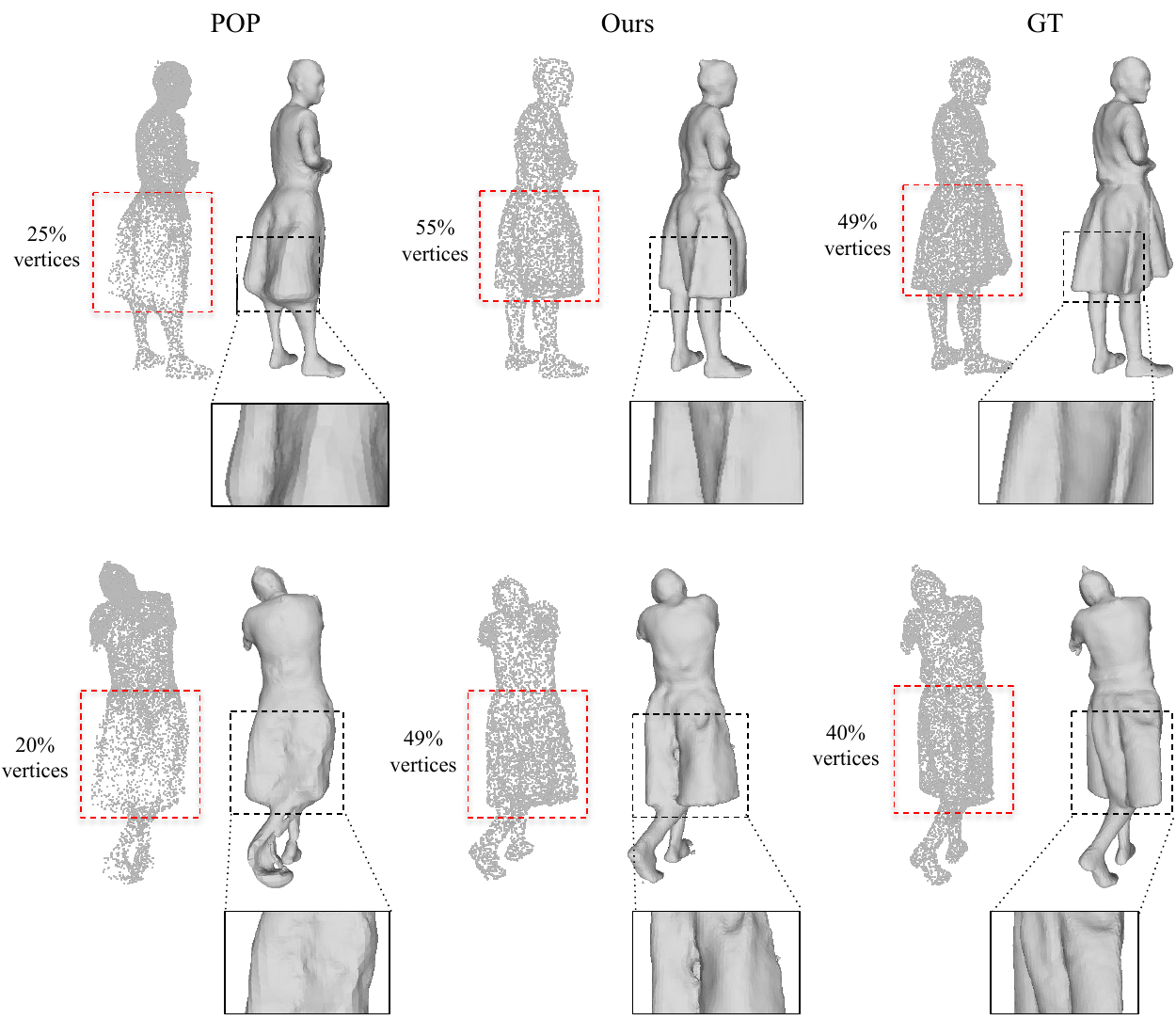}
  \caption{Qualitative comparison to POP showing results for loose clothing on unseen poses. We show both point clouds and their meshified versions for depicting point density and clothing deformation respectively. We additionally show the percentage of vertices occupying the loose clothing region (skirt). Due to modeling the clothing on top of a template model such as SMPL, the points from POP in the skirt region are too sparse to model any significant deformations. This is due to the points having a hard association with the nearest body part. Our method produces points much more consistently distributed across the body and clothing, thereby exhibiting realistic pose-dependent clothing deformations. Zoomed-in regions emphasize the most significant clothing deformations. }
  \label{fig:pop_supp}
\end{figure*}

\FloatBarrier

\section{Extended Results and Discussion}
\label{sec:qualitative_ext}
Here, we conduct further analysis of our results in the main paper. We compare with SoTA point-based method POP and also show how a posed SMPL mesh can be used to refine the obtained avatar from PocoLoco. Finally, we also provide more ablation studies to give an insight into the selection of our architecture.

\clearpage
\twocolumn

%
%

\twocolumn[{%
\renewcommand\twocolumn[1][]{#1}%
\includegraphics[width=\linewidth]{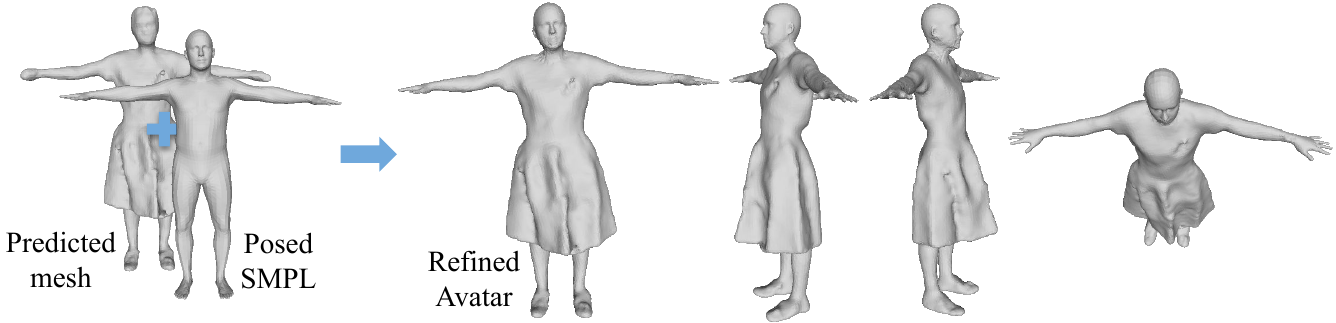}
\captionof{figure}{
     Using SMPL as a post-processing step helps in recovering face and the more difficult to obtain hand details. 
}
\label{fig:smpl_refinement}
\vspace{5mm}
}]

\subsection{Comparison with POP} 
\vspace{-1mm}
\label{sec:pop_sec}

POP shows quantitative results for multi-subject training. For a fair comparison, we train and evaluate POP in a subject-specific manner, similar to how we report results for PocoLoco. 

\noindent \textbf{Points sparsity.} We show an in-depth comparison with POP here. The illustration in \cref{fig:pop_supp} reveals that points from POP exhibit sparsity in the skirt region, detecting fewer clothing deformations. In contrast, PocoLoco generates points evenly distributed across the body and clothing, enabling the recognition of clothing deformations. We quantify this by counting the number of vertices in the skirt region. While POP allocates approximately 25\% of its points to the skirt region, PocoLoco assigns approximately 50\% of its points to this area, contributing to the detection of significant deformations in the skirt region.

\begin{figure*}[!ht]
	\centering
	\includegraphics[width=.95
 \linewidth]{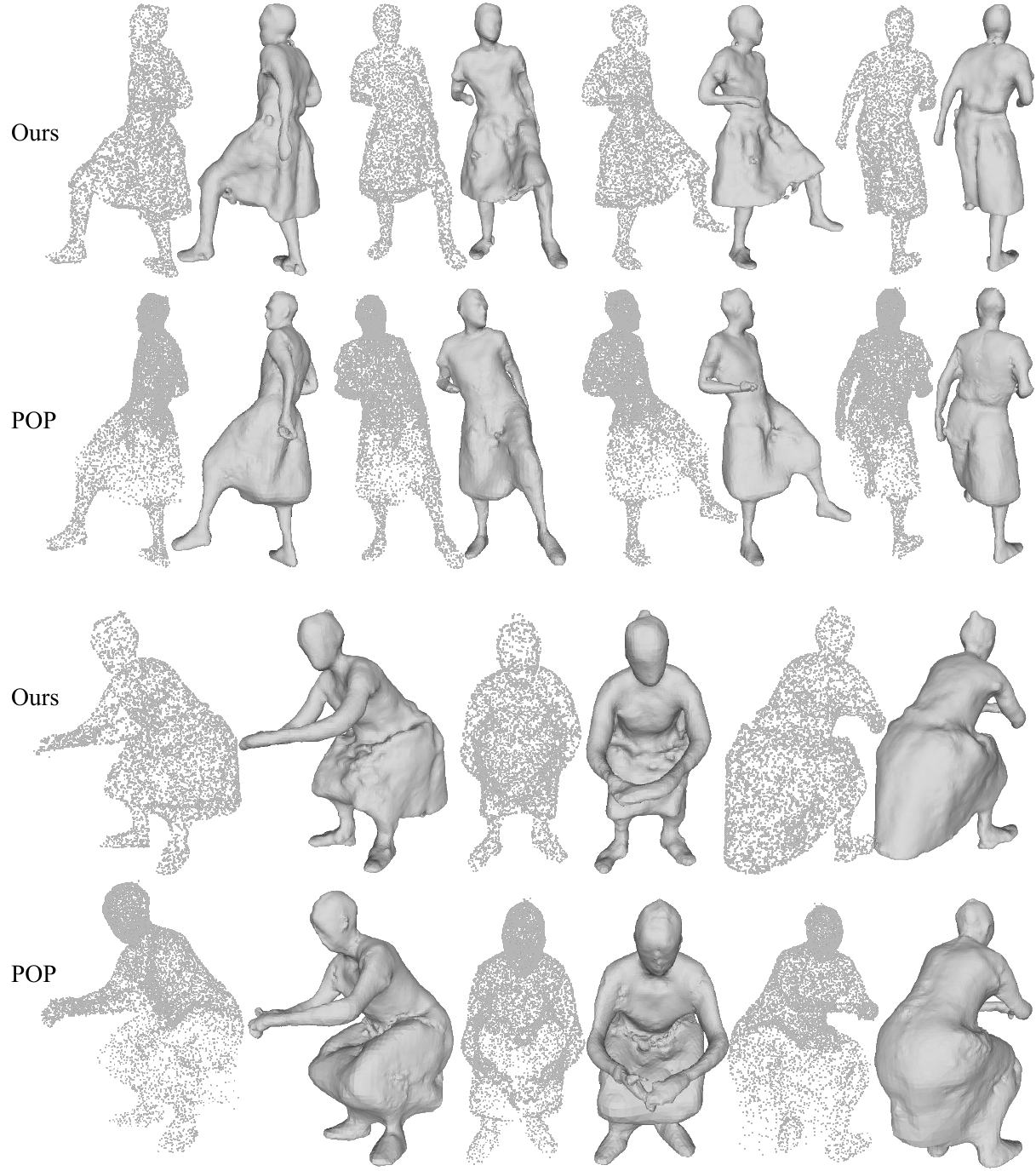}
	\caption
	{
	    Qualitative comparison on top 10\% most difficult poses in the loose subject of DynaCap dataset. POP obtains 7.2 cm CD while PocoLoco obtains 5.5 cm CD.
	}
	\label{fig:oodposes}
\end{figure*}

\vspace{1mm}
\noindent \textbf{Performance on most difficult poses.} \cref{fig:oodposes} shows a performance comparison on the top 10\% most difficult poses in the test set of the loose clothing subject in DynaCap. For each scan in the test set, we find the closest appearing sample in the train set and measure the CD. The samples in the test set with the highest CD are considered the most difficult poses as they do not appear in the train set. We pick the top 10\% of such samples and show a quantitative comparison in \cref{fig:oodposes}.

\vspace{1mm}
\noindent \textbf{Performance on LOOSE dataset.} Lastly, we evaluate POP on our LOOSE dataset. POP achieves a CD of 1.95 cm (vs Ours 2.87 cm) on Subject 1 and 2.86 cm (vs Ours 3.63 cm) on Subject 2.

\noindent \textbf{Reproducibility.} Prior arts like POP \cite{POP:ICCV:2021} require the registration of an SMPL \cite{SMPL:2015} body model to the 3D reconstructions before the training process, but unfortunately do not provide the code to obtain these in their public repositories. As our model is purely learning-based and does not require any prior registration of a scanned template or human body model, it will be completely reproducible on any other data given our released codebase.

%

\subsection{SMPL based refinement} 
\vspace{-1mm}
\label{sec:smpl_sec}

While previous approaches \cite{POP:ICCV:2021} utilize a template such as SMPL to constrain the space of deformations using Linear Blend Skinning, the resulting clothed meshes suffer from artifacts such as a tear in the skirt region due to points sparsity in modeling loose clothes. Our diffusion-based architecture attends to the uniform distribution of points in the loose clothing region thereby modeling clothing deformations. However, it may lose out on prior information such as facial and hand geometry available to template-based methods. We propose to mitigate this problem via a post-refinement step. Once we obtain the pose-conditioned point clouds from the inference pipeline, we fit a SMPL template to this test-time predicted unseen point cloud. Following this, we extract the head and hand regions from the SMPL template and replace them with our predictions to obtain a higher-quality posed avatar. \cref{fig:smpl_refinement} illustrates the results obtained using this approach. This is similar to how ECON \cite{xiu2023econ} proposes an optional stage to obtain the final mesh.

\begin{figure}[t]
	\centering
	\includegraphics[width=.95
 \linewidth]{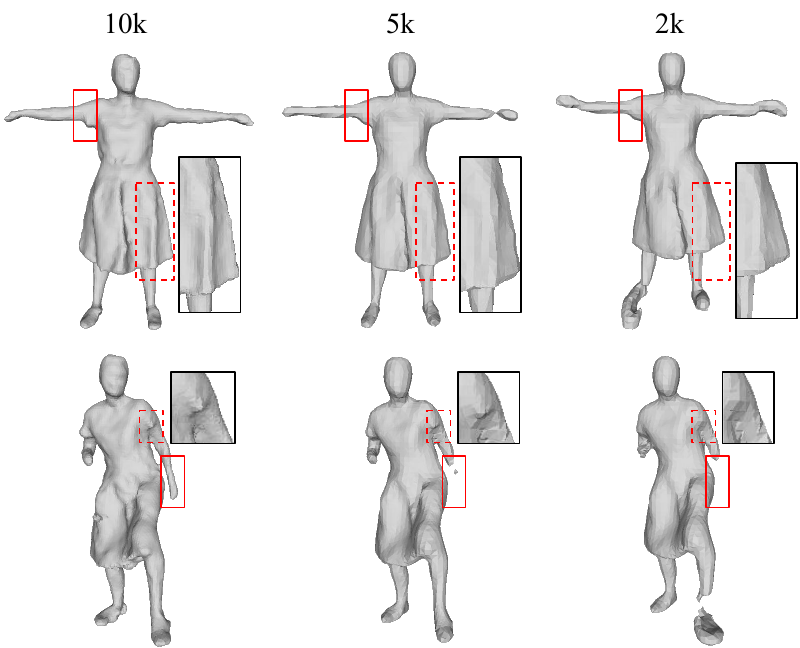}
	\caption
	{
	    Visualization of modeling clothing deformation as a function of the number of points. More points yield better representation capability.
	    We obtain a CD of 5.44 cm for 2k, 4.62 cm for 5k, and 4.22 cm for 10k points.
	}
	\label{fig:points}
\end{figure}

\subsection{Ablation studies} 
\vspace{-1mm}
\label{sec:points_sec}

\noindent \textbf{Importance of the number of points.} 
We perform an ablation study to measure the performance as a function of the number of points used for training our method. \cref{fig:points} shows that more points yield better performance. In the top row, the model starts to lose out on details in the shirt's sleeve region as the number of points reduces. We see a similar effect in the skirt region as well. In the bottom row, points on the left hand become sparse with respect to the overall points. As the hand is the thinnest part of the body, we see a part of it does not have enough points to get the faces. 

\begin{figure}[t]
	\centering
	\includegraphics[width=.95
 \linewidth]{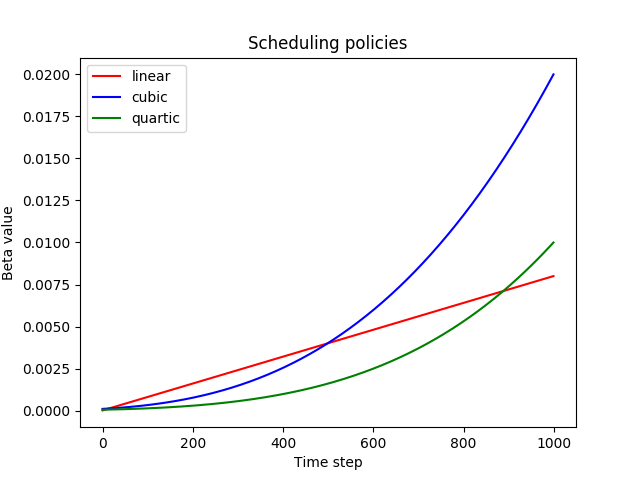}
	\caption
	{
	    We depict the effect of sampling betas based on different noise scheduling policies used in our diffusion model. We choose quartic as it adds noise slowly at the beginning thereby retaining more details for a longer duration than the linear schedule. Best viewed in color.
	}
	\label{fig:scheduling_graph}
\end{figure}

\vspace{1mm}
\noindent \textbf{Scheduling policy.} We propose the Quartic scheduling policy for our diffusion-based architecture which helps to recover more details compared to the Linear scheduling algorithm. As \cref{fig:scheduling_graph} illustrates, the Quartic policy uses smaller beta values at the beginning and gradually uses higher values to convert the point clouds to a Gaussian distribution noise. This implies details such as clothing deformations are retained for more time steps during the forward diffusion step, and the coarse body shape is converted to noise in subsequent steps. During the reverse diffusion process, we benefit from recovering the coarse body shape early on so that the model can utilize more time steps towards recovering clothing deformations. We also experiment with the Cubic scheduling policy. However, beta values obtained using the quartic schedule work best for us.

\noindent \textbf{Training time.} The training time varies for PocoLoco from 4 (19500 frames) to 6 days (33500 frames) with 8 A100 GPUs. However, as we use a Transformer architecture we note that by using FlashAttention, the training time effectively reduces by 4x. Though we do not conduct a quantitative evaluation, we see no visible artifacts in the generated predictions using FlashAttention. Inference time is 80s per sample which can again be reduced using FlashAttention.
\vspace{1mm}

\begin{figure}[t]
	\centering
	\includegraphics[width=.95\linewidth]{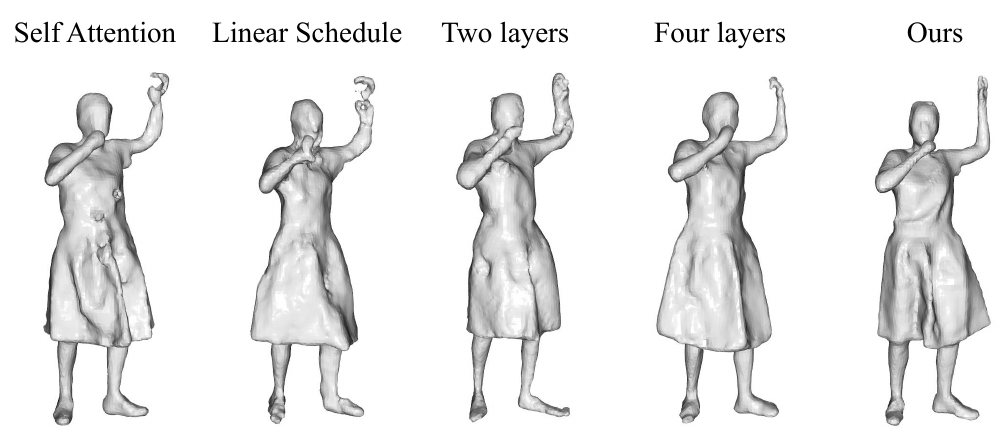}
	\caption
	{
	    Ablation study to show the effectiveness of proposed components (using only self-attention, a linear noise schedule, or self and cross-attention with two and four layers) against the full model (rightmost). We note the importance of each of our design choices for the quality of the final generation results.
	}
	\label{fig:ablation}
\end{figure}

\noindent \textbf{Cross, Self, Cross+Self Attention.} We show in \cref{fig:ablation} the efficacy of using only self, only cross, and a combination of self and cross attention in the proposed Transformer architecture. Though self-attention works reasonably well, it takes a longer time to converge. Cross attention on the other hand does not converge after a long training time but can reasonably model the pose. A combination of self and cross-attention brings the best of both worlds by attending to the conditioned pose and converging to the target point cloud faster.

%
%

\subsection{Comparison with SkiRT} 
\vspace{-1mm}
\label{sec:skirt_sec}

\cite{skirt2022} extend POP to predict blended skinning weights for each point. Besides the data used in training POP, one needs to have some \textit{extra parameters} (as mentioned on the project page) to train SkiRT on a custom dataset that the authors have not yet discussed details about. It is thus not possible to reproduce the results from SkiRT on our dataset. We do not train our model on the ReSynth dataset as it only has about 1000 frames per subject which is insufficient to train a diffusion model. We refer the reader to the supplementary section S3.1 in \cite{skirt2022} where the authors note that the points produced by SkiRT can still be visibly sparser than on other body parts. PocoLoco on the other hand does not suffer from such a problem.

\subsection{Application: Pose Editing} 
\vspace{-1mm}
\label{sec:editing_sec}
We can achieve pose editing in two ways. First, consider the problem of point cloud (PC) completion. For a partial PC $\in R^{(N-K)\times3}$ with K missing points, we train our model to reconstruct the target PC $\in R^{N\times3}$ by denoising K points sampled from Gaussian distribution. Similarly, we remove the points in the area of pose difference and proceed to reconstruct w.r.t. the conditioned target pose. The second approach is to add $t=100$ steps of noise to the source pose PC and condition this with the target pose to get the pose edited result. We note that this works well for minor pose changes and may result in slight shape change for a major change in pose.

\begin{figure}
	\includegraphics[width=2.95in ]{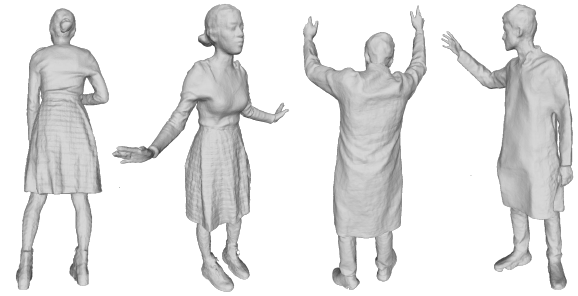}
        \vspace{-2mm}
	\caption
	{LOOSE dataset with 49K points, showing folds and wrinkles in striped skirts and long shirts. Best viewed zoomed in.}
	\vspace{-4mm}
	\label{fig:hqdataset_figure}
\end{figure}

\subsection{LOOSE Dataset details} 
\vspace{-1mm}
\label{sec:loosedetails_sec}
We follow a process similar to DynaCap~\cite{habermann2021real} for a convenient benchmarking method. To create the LOOSE dataset, we begin by scanning the actor in a T-pose with a 3D scanner. We then use commercial multi-view stereo reconstruction software, PhotoScan (http://www.agisoft.com) to generate the 3D mesh. This mesh is manually rigged to a skeleton. Additionally, we track human motions using a multi-view markerless motion capture system, TheCaptury (http://www.thecaptury.com/). \cref{fig:hqdataset_figure} shows our dataset with high-quality geometric details using 49K points.

\section{Limitations and future work} 
\vspace{-1mm}
\label{sec:limit}

While our method is adept at recovering loose clothing deformations, modeling fine details such as facial and hand geometry is difficult. Prior arts benefit from this by using a SMPL template as a prior which helps them retain these details. We see this in \cref{fig:pop_supp} where POP models face geometry better than PocoLoco, albeit different from the GT. We propose a way to mitigate this in the post-processing step of our scans where we fit the posed SMPL meshes. On another note, we are limited to using 10k points due to our computational heavy transformer architecture. As we show in \cref{fig:points}, using more helps recover more details. We leave the design of a more efficient architecture as a future work. Furthermore, the method may fail for extreme unseen poses.

Finally, since our method does not consider temporal consistency, a motion sequence extracted as a sequence of poses may exhibit noticeable changes in deformations for similar poses, leading to animation that lacks smoothness. We regard addressing this as a potential avenue for future research.